\documentclass{article}
\usepackage{spconf,amsmath,graphicx}
\usepackage{CJKutf8}
\usepackage{multirow}
\usepackage{booktabs}
\usepackage{amssymb}

\title{KNOWLEDGE DISTILLATION FROM LANGUAGE MODEL TO ACOUSTIC MODEL:  \\ A HIERARCHICAL MULTI-TASK LEARNING APPROACH}
\name{Mun-Hak Lee and Joon-Hyuk Chang\thanks{Thanks to XYZ agency for funding.}}
\address{Department of Electronics Engineering\\ Hanyang University, Seoul, Republic of Korea}
\begin{document}
\begin{CJK}{UTF8}{mj}

%
\maketitle
\begin{abstract}

The remarkable performance of the pre-trained language model (LM) using self-supervised learning has led to a major paradigm shift in the study of natural language processing. In line with these changes, leveraging the performance of speech recognition systems with massive deep learning-based LMs is a major topic of speech recognition research. Among the various methods of applying LMs to speech recognition systems, in this paper, we focus on a cross-modal knowledge distillation method that transfers knowledge between two types of deep neural networks with different modalities. We propose an acoustic model structure with multiple auxiliary output layers for cross-modal distillation and demonstrate that the proposed method effectively compensates for the shortcomings of the existing label-interpolation-based distillation method. In addition, we extend the proposed method to a hierarchical distillation method using LMs trained in different units (senones, monophones, and subwords) and reveal the effectiveness of the hierarchical distillation method through an ablation study. 

\end{abstract}

\section{Introduction}

Humans transmit language information through voice signals. Automatic speech recognition (ASR) is a technology that extracts language information in the spoken voice signal, facilitating natural communication between humans and machines. The acoustic model (AM) is a major component of the speech recognition system and plays a role in estimating meaningful recognition units (words, subwords, phonemes, letters, etc.) from a given speech signal. The AM learns the correlation between speech and text using a human-tagged speech-transcription paired dataset. However, if sufficient training data are not secured or a domain is mismatched between the training and test datasets, the performance of the speech recognition system is degraded, a considerable obstacle to the stable operation of the speech recognition system. One way to overcome this limitation is to use a language model (LM). The LM learns the conditional probability distribution of the word sequences using a large unannotated corpus and helps the speech recognition system trained only with limited paired data to model unseen words or sentences well. Recently, the deep learning-based LM \cite{fnn_lm, rnn_lm} has successfully overcome the long-term dependency problem and sparsity problem, which are shortcomings of the statistical LMs, and large-scale LMs pre-trained with self-supervised learning frameworks \cite{bert} have achieved remarkable success in the field of natural language processing.


Therefore, making full use of the remarkable performance of deep learning-based LMs in speech recognition systems is a major research direction for speech recognition technology. From shallow fusion \cite{differentialble_beam_search} to re-scoring methods \cite{lm_rescoring, lattice_rescoring}, various studies have been conducted to apply deep learning-based LMs to ASR systems. Among these various research topics, we focus on methods using knowledge distillation \cite{knowledge_distillation, lst, lst_bert}. The knowledge distillation method has advantages in that it does not increase the computation in the decoding process of the speech recognition system and can be combined with shallow fusion or re-scoring methods to further enhance recognition performance.


In this paper, we propose a novel knowledge distillation method based on multi-task learning and apply the proposed method to the AM of the hidden Markov model (HMM)-based hybrid speech recognition system \cite{hyb_asr} and the attention-based sequence-to-sequence (seq2seq) speech recognition model \cite{seq2seq_asr}. The proposed method outperforms the existing label-interpolation-based knowledge distillation in the seq2seq speech recognition system \cite{lst, lst_bert} and has the advantage of operating more stably at various hyper-parameter settings. Unlike the original knowledge distillation, the multi-task learning-based approach can transfer between neural networks with different output units (such as AM/LM in HMM-based speech recognition systems). Using these features, we propose a hierarchical distillation method that transfers the knowledge of multiple LMs with different output units to an HMM-based AM, effectively enhancing the classification performance of the AM. To the best of our knowledge, this is the first attempt to conduct knowledge distillation from pre-trained LMs to HMM-based AMs.

\begin{figure}[t]
  \centering
  \includegraphics[scale=0.7]{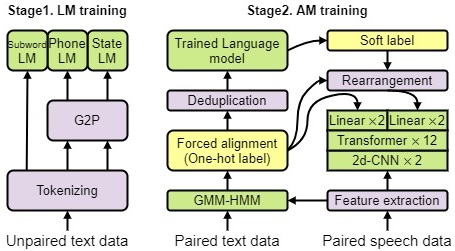}
  \caption{
  Schematic diagram of hierarchical distillation method applied to DNN-HMM-based hybrid speech recognition system. In stage 1, we train LMs with different output units (senones, monophones, and subwords) through manual tokenization and G2P processes, and in stage 2, we transfer the knowledge of the trained LMs to a single acoustic model.}
\end{figure}

\begin{figure}[t]
  \centering
  \includegraphics[scale=0.7]{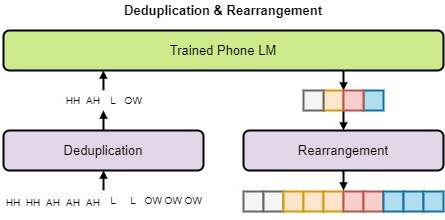}
  \caption{
  Deduplication: Several consecutive overlapping labels are distributed in the forced alignment generated through the Gaussian Mixture-Hidden Markov Model-based speech recognition system (GMM-HMM-based ASR). We remove the overlapped labels through the deduplication algorithm (this algorithm also includes matching the units of forced alignment and LM outputs). Rearrangement: The rearrangement algorithm aligns the list of posterior distributions of the LM one-to-one with the forced alignment before the deduplication process.}
\end{figure}

\section{Background}
\subsection{Knowledge distillation}

A model ensemble of multiple models exhibits higher generalization performance than a single model \cite{ensemble}. \cite{knowledge_distillation} proposed a knowledge distillation method that can transfer the performance of a massive ensemble model to a lighter model. In \cite{knowledge_distillation}, the knowledge of the teacher model is assumed to be compressed in its output distribution, and the student model is trained to mimic the output of the teacher model. Therefore, this method is also called the student-teacher method. The learning loss of the student network for knowledge distillation is as follows:
\begin{equation}\label{eq1}
  \mathcal L_{KD} =  \sum_{i=1}^K 	 \mathcal D( {\exp (v_{i}/T) \over \Sigma^{K}_{j} \exp (v_{j}/T)}, {\exp u_{i} \over \Sigma^{K}_{j} \exp u_{j}} ),
\end{equation}
where $\mathcal D$ is a distance measure, ($v, u$) denote the unnormalized output of the (teacher, student) models, $K$ represents the number of classes, and $T$ is the temperature value.

\subsection{Cross-modal distillation with the language model}
Cross-modal distillation is a useful technique to alleviate the chronic data shortage problem of deep learning in that it can transfer knowledge learned with rich modalities to other modalities provided with only limited labeled data \cite{speech2text}. In this chapter, we briefly introduce the learn spelling from teachers (LST) method, a cross-modal distillation method proposed by \cite{lst}. The LST method transfers the knowledge of the teacher LM to the seq2seq AM sharing the same output unit using the loss function below:
\begin{equation}\label{eq1}
  \mathcal L_{LST} = \lambda \mathcal L_{CE} + (1-\lambda) \mathcal L_{KD},
\end{equation}
where $\mathcal L_{CE}$ is the cross-entropy loss between student network output and true label and $\lambda \in [0,1]$ denotes the tuning factor. If the Kullback–Leibler divergence (KLD) is used as a distance measure for $\mathcal L_{KD}$, the above equation is equivalent to calculating the KLD between the new target distribution generated by interpolating two labels (soft label and hard label) and the output of the student network, where the soft label is the normalized teacher model output. The following equation can express this: 
\begin{align}\label{eq1}
    \mathcal L_{LST} = - \sum_{i=1}^K \hat{P}_i \log{ {\exp u_{i} \over \Sigma^{K}_{j} \exp u_{j}}}, \\
    \hat{P}_i = \lambda Y_i + (1-\lambda) {\exp (v_{i}/T) \over \Sigma^{K}_{j} \exp (v_{j}/T)},\notag
\end{align}
where $Y$ is a one-hot encoded true label distribution. Therefore, this method is referred to as label interpolation-based knowledge distillation \cite{lst}. This method concisely integrates supervised learning loss and knowledge distillation loss into one, but it has clear limitations. First, for label interpolation, the teacher and the student models must share the same output unit. This constraint limits the algorithm scalability; for example, it is impossible to distill knowledge between neural nets with different outputs, such as the LM and AM of an HMM-based speech recognition system. Second, label smoothing is performed in the label interpolation process, which can cause an under-confidence problem in the second and third best classes of the network output. We deal with these issues in more detail in the appendix \cite{appendix}. Third, the LST method has two hyper-parameters ($T, \lambda$) to adjust the soft label sharpness and interpolation ratio, and the network performance responds sensitively according to the setting of the two tuning factors. Thus, when training a network with a new data set, the method must go through a hyper-parameter searching process that takes substantial time and computation. The proposed method separates the two tasks of supervised learning and knowledge distillation, solving the problems of the interpolation-based knowledge distillation method.


\section{Proposed method}
\subsection{A multi-task learning approach for knowledge distillation}

The proposed multi-task learning-based knowledge distillation method is designed to compensate for the shortcomings of the label interpolation-based knowledge distillation method. As illustrated in Figure 1, the proposed AM consists of shared encoding layers (2D-convolution neural network (CNN) + transformer), two linear layers for supervised learning, and two auxiliary linear layers for knowledge distillation. Therefore, the proposed model has two outputs, and each output approximates a different target distribution (hard and soft labels):
\begin{multline}
    \mathcal L_{proposed} = -\sum_{i=1}^K ( \lambda Y_i \log {  {\exp u^{SL}_{i} \over \Sigma^{K}_{j} \exp u^{SL}_{j}} } \\
    + (1-\lambda) {\exp (v_{i}/T) \over \Sigma^{K}_{j} \exp (v_{j}/T)} \log{ {\exp u^{KD}_{i} \over \Sigma^{K}_{j} \exp u^{KD}_{j}}})
\end{multline}
where $u^{KD}$ is the unnormalized output for knowledge distillation and $u^{SL}$ denotes the unnormalized output for supervised learning. We summarize the advantages of the proposed multi-task learning-based knowledge distillation method as follows:
\begin{enumerate}
    \item The LST method has the constraint that the teacher and student share the same output unit. The proposed method is free from these constraints and is able to use pre-trained LM with various output units.
    \item It is possible to transfer the knowledge of LMs with different output units to one AM, and knowledge distillation using various intermediate level units improves AM performance.
    \item The LST method has two hyper-parameters ($T, \lambda$), and the recognition performance varies greatly according to the change in hyper-parameters. The proposed method works stably in most hyper-parameter settings.
    \item The LST model is trained by targeting the smoothed label generated through label interpolation, amplifying the calibration error for the second and third best classes of the AM. In addition, the calibration error of the AM leads to the deterioration of beam search decoding performance \cite{better_decoding, mypaper}. The proposed method solves this problem by separating knowledge distillation and supervised learning tasks.

\end{enumerate}

\subsection{Hierarchical knowledge distillation through multi-task learning}

A language has a hierarchical structure composed of sentences, words, and characters. \cite{hierar_ctc, hierar_seq2seq} used this hierarchical structure to improve the classification performance of AMs. Taking advantage of the multi-task learning approach, we transfer the knowledge of multiple LMs with different output units (mono phones and subwords) to a single deep neural network (DNN) HMM-based AM with a senone (decision-tree-based tri-phone) as the output unit. Two types of empirical techniques are needed to proceed with this hierarchical distillation. The first is a manual tokenizing and grapheme to phoneme (G2P) algorithm. Through this, we can replace the unannotated corpus with a smaller unit of interest, and the forced alignment can be replaced with the desired recognition unit through inverse transformation. The second is to align the LM posterior distribution and speech feature arrangement so that the DNN can be trained. As depicted in Figure 2, we create a frame-wise LM posterior from forced alignment through two steps (deduplication and rearrangement) and transfer the LM knowledge to a DNN-HMM-based AM using this method.

\section{Experimental setup}
\subsection{Datasets}

We conducted the experiments using the LibriSpeech dataset. The LibriSpeech dataset consists of 460h of the clean training set, 500h of the more challenging training set, and separate validation and test sets. In all experiments, we trained the ASR system using only 100h of the training set. The test set of LibriSpeech is divided into ``clean" and ``other". We used the text corpus with about 40 million sentences from the LibriSpeech dataset for all LM training.

\subsection{Model architecture}

For the end-to-end (E2E) ASR experiment, we used an attention-based seq2seq structure. The encoder and decoder of the Seq2seq network consist of a transformer (12 and 6 layers, respectively), and two 2D-CNN-based subsampling layers are added to the encoder input. For E2E ASR, we used 5,000 subwords \cite{sentencepiece} as output units for both the LM and seq2seq AM. For LM training, we used a simple network structure in which four layers of long short-term memory were stacked after the embedding layer.


As an acoustic model of the DNN-HMM-based hybrid speech recognition system, we used a DNN structure in which 12 transformer layers and two linear output layers were stacked on top of a 2D-CNN of two layers. We created forced-alignments for DNN training with the trained GMM-HMM model, and the forced-alignment consists of 4,160 senones. We additionally used a tri-gram-based external LM in the decoding process of an experiment using a hybrid speech recognition system \cite{srilm}. The hybrid and Seq2Seq ASR systems were trained using the 80-dimensional filter bank feature, and spec-augmentation was applied to ensure stable performance \cite{specaug}. The speech recognition systems we used in the experiments are based on the codes of ESPNet and Kaldi \cite{espnet, kaldi}.

\begin{figure}[t]
  \centering
  \includegraphics[scale=0.2]{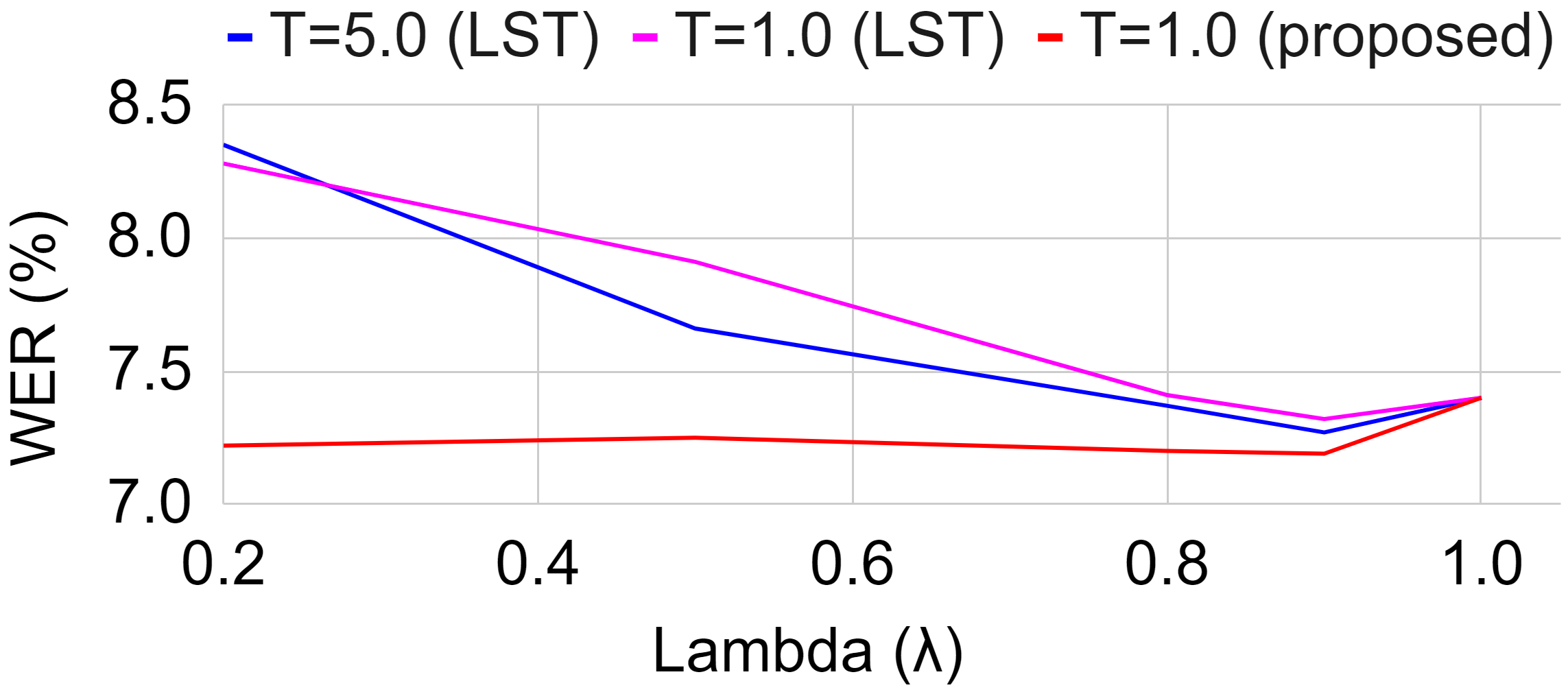}
  \caption{
  Effect of varying the interpolation factor $\lambda$ on the LibriSpeech test-clean. We quantitatively compare the interpolation-based knowledge distillation method (LST) with the proposed method.
  }
\end{figure}

\section{Results and analysis}
\subsection{Learn spelling from teacher vs the proposed method}

This chapter compares the differences between the interpolation-based knowledge distillation method and the proposed multitask learning approach. We used a seq2seq model with similar settings to \cite{lst} for the experiments, and let the both teacher network (LM) and student network (seq2seq) share the same output unit (5,000 subwords). As the first experiment, we diversified the interpolation factor ($\lambda$) in the loss function of (3). We fixed the $T$ value to 5.0, and the experimental results are presented at the top of Figure 3. Second, we added two task-specific layers to the seq2seq model decoder and performed knowledge distillation using the loss function of (4). We experimented by changing the value of $\lambda$ while fixing the value of $T$ at 1.0. The experiment confirmed that the method of (3) is highly dependent on the $\lambda$ value, and in some experiments, it has lower performance than the baseline model trained only with supervised learning. However, the proposed loss function, (4), exhibited more stable performance for various parameter settings and better performance than the interpolation method in all experiments.

\subsection{Hierarchical knowledge distillation through multi-task learning}

In this chapter, we experimented with a hierarchical distillation method that transfers the knowledge of several LMs with different outputs to a single AM using multiple auxiliary layers. We trained three LMs of the same structure with different output units. The first LM shares the same output unit (4,160 senones) as the AM. The second LM was trained with 41 mono-phones (phones) as an output unit, and the last LM was trained with 5,000 subwords \cite{sentencepiece} as an output unit. In the knowledge distillation experiment using only one LM, the experiment using the subword unit demonstrated the highest relative performance gain of about 7.5\%. In the hierarchical distillation experiment using multiple LMs, the experiment using all three LMs had the highest relative performance gain of 9\%. We list these results in Table 1. These experimental results are consistent with the previous study in that adding an auxiliary task using the hierarchical structure of speech improves the classification performance of the AMs \cite{hierar_ctc, hierar_seq2seq}.

\begin{table}[]
\centering
\caption{\label{tab:table-1} 
AM classification accuracy (ACC) and word error rate (WER) for hybrid ASR system. We show the results of an ablation study using three types of LMs with different output units.
}

\begin{tabular}{ccccc}
\toprule
\multicolumn{3}{c}{LM unit} & \multirow{2}{*}{\begin{tabular}[c]{@{}c@{}}ACC\\ (\%)\end{tabular}} & \multirow{2}{*}{\begin{tabular}[c]{@{}c@{}}WER\\ (\%)\end{tabular}} \\
senone  & phone  & subword  &            &         \\ \hline
 - & - & - & 83.23            & 7.98    \\
$\checkmark$ & - & - & 84.26            & 7.85    \\
- & $\checkmark$ & - & 84.41            & 7.51    \\
- & - & $\checkmark$  & 84.60          & 7.38       \\
$\checkmark$ & $\checkmark$      & - & 84.44            & 7.43    \\
$\checkmark$ & $\checkmark$      & $\checkmark$  & 84.85          & 7.26       \\ \bottomrule
\end{tabular}
\end{table}

\section{Conclusion}


In this study, we proposed a new acoustic model training method that combines multi-task learning and knowledge distillation. We experimentally demonstrated that the proposed method compensates for the weaknesses of the interpolation-based knowledge distillation method. In addition, we proposed a hierarchical distillation method using the hierarchical structure of speech, reducing the relative error rate of the speech recognition system by 9\%. The knowledge distillation algorithm proposed in this study has a strong advantage: the distributed pre-trained LMs can be used regardless of the output unit if the appropriate manual tokenizing/G2P algorithm is secured. We plan to actively use this advantage to conduct knowledge distillation experiments with larger LMs, such as BERT \cite{bert} in the future.

\vfill\pagebreak

\bibliographystyle{IEEEbib}
\bibliography{strings,refs}

\begin{thebibliography}{10}

\bibitem{fnn_lm}
Yoshua Bengio, R{\'e}jean Ducharme, Pascal Vincent, and Christian Janvin,
\newblock ``A neural probabilistic language model,''
\newblock {\em The journal of machine learning research}, vol. 3, pp.
  1137--1155, 2003.

\bibitem{rnn_lm}
Tomas Mikolov, Martin Karafi{\'a}t, Lukas Burget, Jan Cernock{\`y}, and Sanjeev
  Khudanpur,
\newblock ``Recurrent neural network based language model.,''
\newblock 2010.

\bibitem{bert}
Jacob Devlin, Ming{-}Wei Chang, Kenton Lee, and Kristina Toutanova,
\newblock ``{BERT:} pre-training of deep bidirectional transformers for
  language understanding,''
\newblock {\em CoRR}, vol. abs/1810.04805, 2018.

\bibitem{differentialble_beam_search}
Ronan Collobert, Awni Hannun, and Gabriel Synnaeve,
\newblock ``A fully differentiable beam search decoder,''
\newblock in {\em International Conference on Machine Learning}, 2019.

\bibitem{lm_rescoring}
Martin Sundermeyer, Hermann Ney, and Ralf Schl{\"u}ter,
\newblock ``From feedforward to recurrent lstm neural networks for language
  modeling,''
\newblock {\em IEEE/ACM Transactions on Audio, Speech, and Language
  Processing}, vol. 23, no. 3, pp. 517--529, 2015.

\bibitem{lattice_rescoring}
Xunying Liu, Xie Chen, Yongqiang Wang, Mark~JF Gales, and Philip~C Woodland,
\newblock ``Two efficient lattice rescoring methods using recurrent neural
  network language models,''
\newblock {\em IEEE/ACM Transactions on Audio, Speech, and Language
  Processing}, vol. 24, no. 8, pp. 1438--1449, 2016.

\bibitem{knowledge_distillation}
Geoffrey Hinton, Oriol Vinyals, and Jeff Dean,
\newblock ``Distilling the knowledge in a neural network,''
\newblock {\em Proceedings of NIPS Deep Learning and Representation Learning
  Workshop}, 2015.

\bibitem{lst}
Ye~Bai, Jiangyan Yi, Jianhua Tao, Zhengkun Tian, and Zhengqi Wen,
\newblock ``Learn spelling from teachers: Transferring knowledge from language
  models to sequence-to-sequence speech recognition,''
\newblock {\em INTERSPEECH}, 2019.

\bibitem{lst_bert}
Hayato Futami, Hirofumi Inaguma, Sei Ueno, Masato Mimura, Shinsuke Sakai, and
  Tatsuya Kawahara,
\newblock ``Distilling the knowledge of bert for sequence-to-sequence asr,''
\newblock {\em INTERSPEECH}, 2020.

\bibitem{hyb_asr}
Alex Graves, Abdel-rahman Mohamed, and Geoffrey Hinton,
\newblock ``Speech recognition with deep recurrent neural networks,''
\newblock in {\em Proceedings of IEEE International Conference on Acoustics,
  Speech and Signal Processing (ICASSP)}. IEEE, 2013.

\bibitem{seq2seq_asr}
Dzmitry Bahdanau, Jan Chorowski, Dmitriy Serdyuk, Philemon Brakel, and Yoshua
  Bengio,
\newblock ``End-to-end attention-based large vocabulary speech recognition,''
\newblock in {\em Proceedings of IEEE international Conference on Acoustics,
  Speech and Signal Processing (ICASSP)}. IEEE, 2016.

\bibitem{ensemble}
Thomas~G Dietterich,
\newblock ``Ensemble methods in machine learning,''
\newblock in {\em International workshop on multiple classifier systems}.
  Springer, 2000, pp. 1--15.

\bibitem{speech2text}
Won~Ik Cho, Donghyun Kwak, Ji~Won Yoon, and Nam~Soo Kim,
\newblock ``Speech to text adaptation: Towards an efficient cross-modal
  distillation,''
\newblock {\em INTERSPEECH}, 2020.

\bibitem{appendix}
Mun-Hak Lee and Joon-Hyuk Chang,
\newblock ``Knowledge distillation from language model to acoustic model
  (appendix),''
\newblock {\em arXiv preprint}, 2021.

\bibitem{better_decoding}
Jan Chorowski and Navdeep Jaitly,
\newblock ``Towards better decoding and language model integration in sequence
  to sequence models,''
\newblock {\em INTERSPEECH}, 2017.

\bibitem{mypaper}
Mun-Hak Lee and Joon-Hyuk Chang,
\newblock ``Deep neural network calibration for e2e speech recognition
  system,''
\newblock {\em INTERSPEECH}, 2021.

\bibitem{hierar_ctc}
Kalpesh Krishna, Shubham Toshniwal, and Karen Livescu,
\newblock ``Hierarchical multitask learning for ctc-based speech recognition,''
\newblock {\em arXiv preprint arXiv:1807.06234}, 2018.

\bibitem{hierar_seq2seq}
Shubham Toshniwal, Hao Tang, Liang Lu, and Karen Livescu,
\newblock ``Multitask learning with low-level auxiliary tasks for
  encoder-decoder based speech recognition,''
\newblock {\em INTERSPEECH}, 2017.

\bibitem{sentencepiece}
Taku Kudo and John Richardson,
\newblock ``Sentencepiece: A simple and language independent subword tokenizer
  and detokenizer for neural text processing,''
\newblock {\em Proceedings of the Conference on Empirical Methods in Natural
  Language Processing}, 2018.

\bibitem{srilm}
Andreas Stolcke,
\newblock ``Srilm-an extensible language modeling toolkit,''
\newblock in {\em International conference on Spoken Language Processing},
  2002.

\bibitem{specaug}
Daniel~S Park, William Chan, Yu~Zhang, Chung-Cheng Chiu, Barret Zoph, Ekin~D
  Cubuk, and Quoc~V Le,
\newblock ``Specaugment: A simple data augmentation method for automatic speech
  recognition,''
\newblock {\em INTERSPEECH}, 2019.

\bibitem{espnet}
Shinji Watanabe, Takaaki Hori, Shigeki Karita, Tomoki Hayashi, Jiro Nishitoba,
  Yuya Unno, Nelson Enrique~Yalta Soplin, Jahn Heymann, Matthew Wiesner, Nanxin
  Chen, et~al.,
\newblock ``Espnet: End-to-end speech processing toolkit,''
\newblock {\em INTERSPEECH}, 2018.

\bibitem{kaldi}
Daniel Povey et~al.,
\newblock ``The kaldi speech recognition toolkit,''
\newblock in {\em IEEE 2011 workshop on Automatic Speech Recognition and
  Understanding (ASRU)}, 2011.

\end{thebibliography}
\end{CJK}

\end{document}


\begin{CJK}{UTF8}{mj}

\maketitle

\section{Expected calibration error}
The purpose of model calibration is to ensure that the output probability distribution of the network accurately reflects the probability of correct answers for each class. Therefore, if a well calibrated model is used, we can grasp not only the prediction result but also the accuracy of the prediction. A perfectly calibrated network satisfies the following equation.
\begin{equation*}\label{eq2}
    P(Y=i  | \hat{\mathbf{p}}(X) = \mathbf{p} ) = p_i  \quad \text{for} \ \ i = 1,...,k
\end{equation*}
where $\mathbf{p}=(p_1,...,p_k)$, $Y$ is the true label and $\hat{\mathbf{p}}$ is the output probability distribution of the $k$-class classification model.  The most representative calibration measure is the expected calibration error (ECE) in Eq. (1).


 \begin{equation}\label{eq3}
    ECE = \sum_{i=1}^{b} {\left\vert B_i \right\vert \over n} \left\vert acc(B_i) - conf(B_i) \right\vert,
\end{equation}
\begin{align*}\label{eq4}
    acc(B_i) &= {1 \over \left\vert B_i \right\vert} \sum_{m \in B_i}^{}  \textbf{1}(\hat{y}_m = y_m), \\
    conf(B_i) &= {1 \over \left\vert B_i \right\vert} \sum_{m \in B_i}^{}  \hat{p}_m,
\end{align*}
where $b$ is the number of bins and $n$ is the total number of data points. ECE measures the difference between accuracy and confidence (probability value for one best class) per bin. Also, it is also important to determine a suitable binning method for the calibration measurements. For this, we used a method of generating bins by sorting the classification results in mini-batch according to confidence scores. Alignment was performed once for each class and once for each mini-batch. If binning is performed in this way, the variance in the bin is minimized; this helps to identify calibration errors for each confidence value.

\begin{figure*}[h]
  \centering
  \includegraphics[scale=0.27]{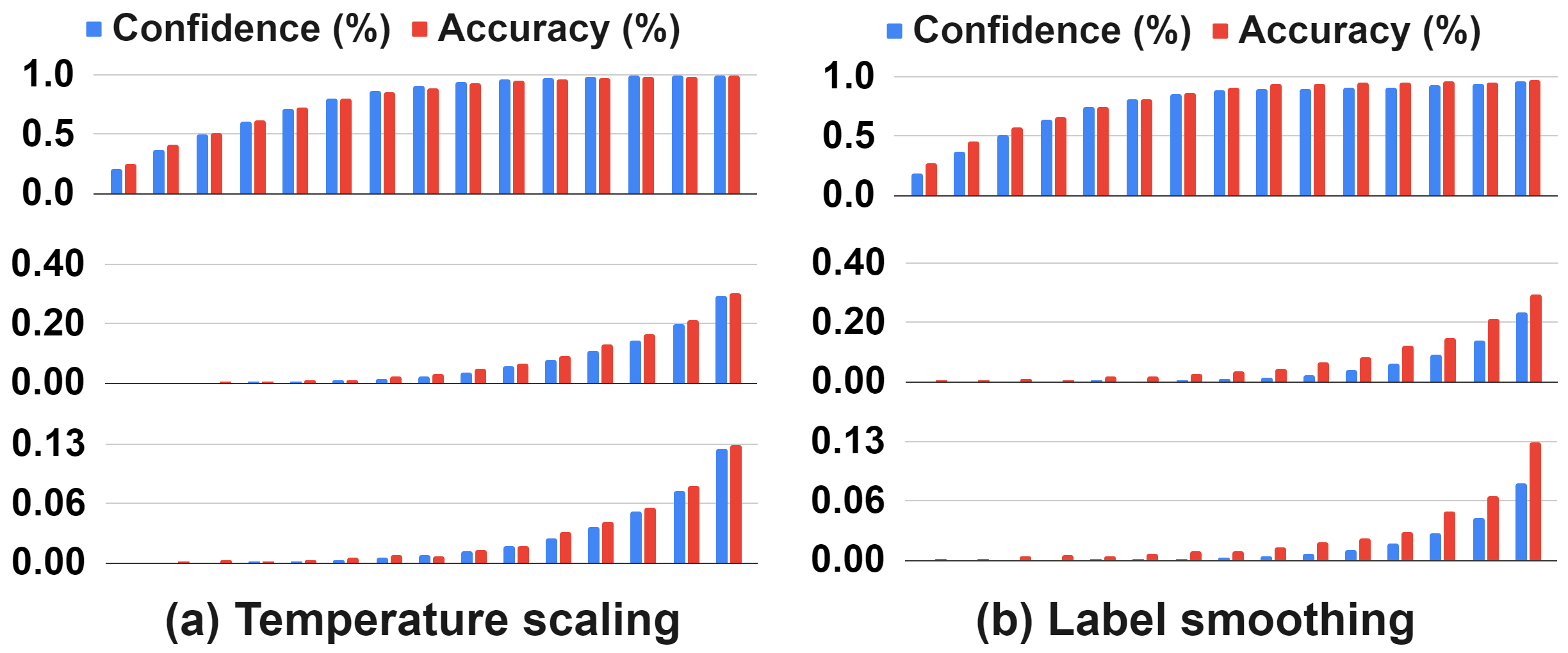}
  \caption{
  Confidence is the posterior probability value for the $N$th best class, and the calibration error is the difference between the average accuracy and the average confidence for each bin \cite{modern_calibration}. (a) A graph comparing the bin-wise averaged confidence and accuracy of top 3 classes of transformer AM calibrated using the temperature scaling method. We can see that the calibration error is relatively small. (b) This is the case calibrated using the label smoothing method, and the label smoothing method only generates a well calibrated posterior probability for the 1 best class (top), but tends to be under-confident for the 2nd and 3rd best (middle, bottom) classes. This is because the label smoothing method gives a small probability value for all trivial classes. This calibration error causes a problem in the beam search decoding stage \cite{better_decoding, mypaper}. We created a graph by dividing the LibriSpeech test-other into 15 bins in total.
  }
\end{figure*}

\section{Calibration methods}

\subsection{Label smoothing}
The method that is often used to solve overconfidence in speech recognition systems is label smoothing. Label smoothing uses the target vector smoothed through the following equation for network training, and prevents the network from generating excessively large output values for one class. Many previous studies have shown that label smoothing is helpful for calibrating neural networks \cite{modern_calibration, better_decoding, when_label_smoothing}.
\begin{equation*}
    y_{smooth}=y_{1hot}- \epsilon(y_{1hot}-{1 \over k} y_{ones}),
\end{equation*}
where $y_{ones}=(1,\ldots,1)$, $y_{1hot}$ are the one-hot target vectors, $k$ is the number of classes, and $\epsilon \in[0,1]$, respectively.

\subsection{Temperature scaling}
Temperature scaling adjusts the sharpness of the output by dividing the unnormalized output of the network by the temperature value ($t$). The $t$ value is trained in a direction that minimizes the loss for the validation set, and the parameters of the classification network are kept fixed during this process \cite{beyond_temperature}. In the decoding process of the speech recognition system, a graph is searched for summing the probability distributions of independently trained modules as follows, and each $t$ value is also independently trained for each module.

\begin{equation}\label{eq7}
  \begin{multlined}
    \hat{W} = \underset{W}{argmax} \ \left\{{1 \over t_{1}} \log(P(X|W)) + {1 \over t_{2}} \log(P(W))\right\},
  \end{multlined}
\end{equation}
where $t_m$ is a scalar value, $W$ is the word sequence and $X$ is a feature.

\section{Label smoothing vs temperature scaling}

Label smoothing and temperature scaling are both widely used calibration methods. Both methods have in common that they adjust the output distribution of the model using a single scalar value (each $T, \epsilon$). However, while label smoothing is applied during model training, temperature scaling is different in that it is a post-hoc calibration method that adjusts the output of an already trained model. Many existing speech recognition papers have shown that label smoothing reduces the calibration error of the neural networks \cite{when_label_smoothing, better_decoding, mypaper}. However, we show in Figure 1 that the label smoothing method reduces the 1-best class calibration error of the network, while amplifying the calibration error for the 2nd and 3rd best classes. The calibration error of these 2nd and 3rd best classes can hinder beam search decoding performance that combines multiple probabilistic models such as language models/acoustic models \cite{better_decoding, mypaper}.


\bibliographystyle{IEEEbib}
\bibliography{strings,refs}
\end{CJK}